\documentclass{Interspeech2024}
\usepackage{utfsym}
\usepackage{colortbl}
\usepackage{subfigure}
\usepackage{caption}
\usepackage{subcaption}
\usepackage{pifont}
\usepackage{multirow}
\usepackage{multicol}
\usepackage{tcolorbox}
\usepackage[compress]{cite}

\interspeechcameraready

\title{On the Effects of Heterogeneous Data Sources on Speech-to-Text\\Foundation Models}
\name[affiliation={1}]{Jinchuan}{Tian}
\name[affiliation={1}]{Yifan}{Peng}
\name[affiliation={1}]{William}{Chen}
\name[affiliation={1}]{Kwanghee}{Choi}
\name[affiliation={1,2}]{Karen}{Livescu}
\name[affiliation={1}]{Shinji}{Watanabe}

\address{
  $^1$Carnegie Mellon University, USA \quad 
  $^2$Toyota Technological Institute at Chicago, USA
}
\email{\{jinchuat,yifanpen,wc4,kwanghec,swatanab\}@andrew.cmu.edu; klivescu@ttic.edu}


\begin{document}

\maketitle
\begin{abstract}
The Open Whisper-style Speech Model (OWSM) series was introduced to achieve full transparency in building advanced speech-to-text (S2T) foundation models. To this end, OWSM models are trained on 25 public speech datasets, which are heterogeneous in multiple ways.
In this study, we advance the OWSM series by introducing OWSM v3.2, which improves on prior models by investigating and addressing the impacts of this data heterogeneity.
Our study begins with a detailed analysis of each dataset, from which we derive two key strategies: data filtering with proxy task to enhance data quality, and the incorporation of punctuation and true-casing using an open large language model (LLM).
With all other configurations staying the same, OWSM v3.2 improves performance over the OWSM v3.1 baseline while using 15\% less training data.  
\end{abstract}

\section{Introduction}
The field of Speech-to-Text (S2T) technology has witnessed remarkable advancements, evolving from simple automatic speech recognition (ASR) \cite{asr_survey2} or speech translation (ST) \cite{st_survey2} applications to complex systems capable of recognizing and translating multiple languages with high accuracy. This evolution has been primarily fueled by the development of large foundation S2T models using massive multilingual corpora \cite{google-usm, seamless, seamlessm4t, pratap2023scaling, whisper}.
A significant milestone in this line of work is the introduction of the Open Whisper-style Speech Model (OWSM), which reproduces Whisper \cite{whisper} and provides better transparency and equal access to such S2T foundation models \cite{owsm, owsm3.1, owsm_ctc}.

To maintain full transparency and reproducibility, the OWSM series relies only on data that is publicly available. However, one challenge with this approach is that no single existing public speech dataset can provide sufficiently massive and diverse data. 
Instead, the OWSM series uses a combination of 25 public datasets from various sources, containing 180k hours and 150 languages.
Prior foundation S2T models were trained on datasets that underwent a standardized pre-processing for all the raw audio \cite{pratap2023scaling, whisper}. However, the datasets we used came from different sources and went through various pre-processing protocols, resulting in heterogeneity that is rarely addressed in the existing literature.
This work is an attempt to address these challenges and subsequently improve model performance by better data consistency.

Our study begins with a detailed analysis of each data involved in training OWSM v3.1 \cite{owsm3.1}, based on which we observe that (1) not all speech-text pairs are well aligned and (2) the text format (especially punctuation and case-sensitivity) of these datasets is not consistent.
First, we conduct proxy tasks to diagnose and remove low-quality data in each dataset, to attempt to ensure that the model learns from more accurately labeled data. 
Second, we do inverse text normalization using Large Language Models (LLMs), ensuring the training text contains punctuation and case-sensitivity uniformly.

Compared with prior data-oriented works, this study is distinctive in the following ways:
(1) Prior works on new datasets \cite{gigaspeech, wenetspeech, takamichi2021jtubespeech, librispeech-corpus} start from unprocessed audio recordings and metadata, aiming to produce a usable dataset for new scenarios by \textit{discarding} low-quality samples. 
Besides, the goal of prior works on active learning and data selection \cite{lu22_interspeech, shef_contrastive, aishell_datadrop, 9383577} is to improve model performance, which is achieved by \textit{expanding} the training corpus using unlabeled or out-of-domain data.
Our work differs from both sides in setups and motivation:
our work starts from a massive, multilingual, but diversely processed data mixture, attempting to achieve performance gain with reduced data volume but higher data quality. 
(2) Conventionally, ASR requires additional post-processing procedures to recover punctuated and capitalized output,
using either weighted finite-state transducers (WFSTs) \cite{zhang21ja_interspeech} or a sequence-to-sequence neural network \cite{sproat2016rnn, paul22_interspeech}. 
Recent works build S2T models with written-form output in an end-to-end manner, but they require extra language model integration \cite{ling2023adapting} and output format design \cite{10389653}.
Compared with prior works, our practice with LLM adoption only revises the training data and requires no post-processing or model modification.

Combining the two aforementioned techniques, this work introduces OWSM v3.2, which advances over the previous OWSM v3.1 \cite{owsm3.1}.
Compared with the OWSM v3.1 baseline, OWSM v3.2 achieves considerable improvement on ST tasks and comparable performance on ASR benchmarks, even with 15\% less training data.
Additionally, evaluation with LLM demonstrates that OWSM v3.2 outputs text that is more aligned with written language with punctuation and case-sensitivity.

\begin{table*}[t]
    \centering
    \caption{Statistics on the OWSM data mixture. Volume includes only the training subset. The data volume can differ from the officially claimed for each dataset due to our data preparation policy. For the \textit{language} column, the 3-character language IDs follow ISO-639-3 standards; digital numbers represent the number of languages for ASR data and the number of translation directions for ST data. License information is based on our collection. Punctuation and Case-Sensitivity specify whether the original text label contains punctuation and is case-sensitive. Dash - means the language is not case-sensitive.
    Long-Form specifies whether the segmentation information is provided to splice short clips into long-form examples.
    }
    \vspace{-6pt}
    \scalebox{0.8}{
    \begin{tabular}{|l|r|r|r|r|c|c|c|c|}
        \hline
         \rowcolor[rgb]{0.9,0.9,0.9} Corpus & Type & Volume (h) & Language & \#Examples & License & Punctuation & Case-Sensitivity & Long-Form \\
         \hline
         aidatatang \cite{aidatatang} & ASR & 140 & zho & 164K & CC-BY-NC-ND-4.0 & \usym{2718} & - &\usym{2718} \\ 
         \rowcolor[rgb]{0.9,0.9,0.9}AISHELL \cite{aishell-corpus} & ASR & 150 & zho & 120K & Apache 2.0 & \usym{2718} & - & \usym{2718} \\
         ami \cite{ami-corpus}  & ASR & 141 &  eng & 24K & CC-BY-4.0 & \usym{2718} & \usym{2718} &\usym{2714}\\
         \rowcolor[rgb]{0.9,0.9,0.9}babel \cite{babel}  & ASR & 2115 & 25 & 318K & - & \usym{2718} & \usym{2718} &\usym{2714}\\
         CommonVoice (CV) \cite{commonvoice} & ASR & 16360 & 104 & 11.8M & CC0-1.0 & \usym{2714} & \usym{2714} &\usym{2718} \\
         \rowcolor[rgb]{0.9,0.9,0.9}CoVoST2  \cite{covost2} & ST & 8550 & 22 & 5.9M & CC-BY-NC 4.0 & \usym{2714} & \usym{2714} &\usym{2718}\\ 
         Fisher Callhome Spanish \cite{fisher-callhome} & ASR & 241 & spa & 36K & - & \usym{2718} & \usym{2718} &\usym{2714}\\
         \rowcolor[rgb]{0.9,0.9,0.9} FLEURS \cite{FLEURS} & ASR & 950 & 102 & 268K & CC-BY-4.0 & \usym{2714} &  \usym{2714} &\usym{2718}\\
         GigaSpeech \cite{gigaspeech} & ASR & 12520 & eng & 2.0M & Apache 2.0 & \usym{2714} & \usym{2718} &\usym{2714} \\
         \rowcolor[rgb]{0.9,0.9,0.9} GigaST \cite{gigast} & ST & 24453 & 2 & 4.0M & CC-BY-NC 4.0 & \usym{2714} & \usym{2714} &\usym{2714} \\
         KsponSpeech  \cite{ksponspeech} & ASR & 960 & kor & 619K &  MIT License & \usym{2718} & - &\usym{2718}\\
         \rowcolor[rgb]{0.9,0.9,0.9} LibriSpeech (LS) \cite{librispeech-corpus} & ASR & 897 & eng & 145K & CC-BY-4.0 & \usym{2718} & \usym{2718}  &\usym{2714} \\
         MagicData (Magic.) \cite{magicdata} & ASR & 711 & zho & 573K & CC-BY-NC-ND-4.0 & \usym{2718} & - &\usym{2718} \\
         \rowcolor[rgb]{0.9,0.9,0.9} Multilingual LibriSpeech (MLS)\cite{pratap2020mls} & ASR & 50670 & 8 & 8.6M & CC-BY-4.0 & \usym{2718} & \usym{2718} &\usym{2714} \\
         MuST-C - ASR part \cite{must-c} & ASR & 2657 & eng & 400K & CC-BY-NC-ND-4.0 & \usym{2714} & \usym{2714} &\usym{2714}\\
         \rowcolor[rgb]{0.9,0.9,0.9}MuST-C - ST part \cite{must-c} & ST & 8163 & 15 & 1.2M & CC-BY-NC-ND-4.0 & \usym{2714} & \usym{2714}  &\usym{2714}\\
         Googlei18n\footnote{Resources 32, 35, 36, 37, 41, 42, 43, 44, 52, 53, 54, 61, 63, 64, 65, 66, 69, 70, 71, 72, 73, 74, 75, 76, 77, 78, 79, and 86 from \url{openslr.org}.} & ASR & 1326 & 21 & 1.0M & CC BY-SA 4.0 & \usym{2718} & \usym{2714} &\usym{2718}\\
         \rowcolor[rgb]{0.9,0.9,0.9}ReazonSpeech \cite{reazonspeech} & ASR & 18864 & jpn & 11.1M & Apache 2.0 & \usym{2714} & -&\usym{2718}  \\
         Russian Open STT \cite{ru-open-stt} & ASR & 4791 & rus & 4.7M & CC-BY-NC & \usym{2718} &\usym{2718} &\usym{2718} \\
         \rowcolor[rgb]{0.9,0.9,0.9}SPGISpeech  \cite{spgispeech}  & ASR & 4999 & eng & 2.0M & - & \usym{2714} & \usym{2714} &\usym{2718} \\
         Fisher SwitchBoard (SWBD) \cite{swbd-corpus} & ASR & 3214 & eng & 498K & - & \usym{2718} & \usym{2718} &\usym{2714}\\
         \rowcolor[rgb]{0.9,0.9,0.9} TEDLIUM3 \cite{tedlium3} & ASR & 472 & eng & 67K & CC-BY-NC-ND 3.0 & \usym{2718} & \usym{2718} &\usym{2714} \\
         VCTK \cite{vctk} & ASR & 25 & eng & 43K & CC-BY-4.0 & \usym{2714} & \usym{2714} &\usym{2718} \\
         \rowcolor[rgb]{0.9,0.9,0.9}VoxForge \cite{voxforge} & ASR & 235 & 8 & 148K & GPL & \usym{2718} &  \usym{2718} &\usym{2718} \\
         VoxPopuli - ASR part \cite{voxpopuli} & ASR & 1702 & 16 & 310K & CC0-1.0 & \usym{2714} & \usym{2714} &\usym{2714} \\
         \rowcolor[rgb]{0.9,0.9,0.9}VoxPopuli - ST part \cite{voxpopuli} & ST & 111 & 40  & 21K & CC0-1.0 & \usym{2714} & \usym{2714} &\usym{2714} \\
         WenetSpeech \cite{wenetspeech} & ASR & 14963 & zho & 2.2M & CC-BY-4.0 & \usym{2718} & - &\usym{2714} \\
         \hline
         \rowcolor[rgb]{0.9,0.9,0.9}Total & & 180396 & 150 & 58.5M & & && \\ 
         \hline
    \end{tabular}}
    \vspace{-10pt}
    \label{tab:stats}
\end{table*}

\section{Methodology}
\subsection{Data Statistics and Analysis}
The OWSM series \cite{owsm, owsm3.1, owsm_ctc} are foundational speech models that support both ASR and ST tasks.
For both ASR and ST, each example in OWSM data can be represented by a tuple $(\mathbf{x}, \mathbf{y}^{\text{src}}, \mathbf{y}^{\text{tgt}}, \mathbf{y}^{\text{prev}})$. 
Here $\mathbf{x}$ stands for speech. 
For ASR, $\mathbf{y}^{\text{src}}$ and $\mathbf{y}^{\text{tgt}}$ are both transcriptions. 
For ST, $\mathbf{y}^{\text{src}}$ and $\mathbf{y}^{\text{tgt}}$ stands for the transcription in the source language and the translation in the target language. 
$\mathbf{y}^{\text{prev}}$ is the previous context of $\mathbf{y}^{\text{tgt}}$.

Table \ref{tab:stats} shows statistics of all 25 datasets adopted in the OWSM v3.1 training, which helps to demonstrate the heterogeneity of the OWSM data mixture.
These datasets primarily differ from each other in types, volumes, and languages.
Although difficult to analyze quantitatively, they come from different acoustic environments, topics, and speaking styles. We expect this diversity to help improve the model's generalizability by increasing the coverage of training data.
However, these datasets also show variance in the following perspectives that may raise issues.

\vspace{-3pt}
\subsubsection{Speech-Text Misalignment}
\vspace{-3pt}
\begin{figure}
    \centering
    \vspace{-7pt}
    \includegraphics{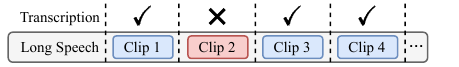}
    \vspace{-10pt}
    \caption{Erroneous long-form example with untranscribed clip}
    \vspace{-10pt}
    \label{fig:clip}
\end{figure}
Some speech and text pairs are not well aligned, at least for the following reasons. 
\textbf{First}, the datasets are built with varying labeling methods.
While some small-scale datasets undergo stringent quality control \cite{aishell-corpus, FLEURS}, massive datasets  \cite{gigaspeech, wenetspeech, commonvoice} often rely on lesser-quality or crowd-sourced transcriptions and undergo automated data adjustments.
These different labeling methods can thus lead to different levels of speech-text misalignment.
The \textbf{second} issue is raised by the \textit{untranscribed clips}.
Unlike the conventional S2T models that mainly work on short clips, the OWSM series is designed to leverage long-form speech context when available.
As shown in Fig.\ref{fig:clip}, within a long speech recording, the speech of a long-form example starts from the first clip and ends at the last, but there can be untranscribed clips in the middle, which leads to a misalignment between the spliced speech and text labels \cite{fox2023updated}.
These ill-aligned examples teach the model to wrongly ignore random intervals of the long speech input during inference, and can subsequently increase deletion errors.
This issue is mainly observed in LibriSpeech, GigaSpeech, and WenetSpeech.


\subsubsection{Inconsistency in Punctuation and Case-Sensitivity}
Conventionally, ASR models are evaluated without punctuation and case-sensitivity, and a large portion of ASR corpora only provide fully normalized transcription\footnote{This issue is less observed in ST datasets: by convention, the ST evaluation includes punctuation and case-sensitivity. All our ST data originally contain punctuation and case-sensitivity.}.
This normalization contrasts with the needs of S2T foundation models like OWSM series, which aims to produce outputs that include such textual features to enhance readability and coherence.
As shown in Table \ref{tab:stats}, previous OWSM series are trained with a data mixture that some datasets contain punctuation and case-sensitivity while others do not, which leads to unpredictable behavior in terms of the output text format.

\begin{table*}
  \caption{
  The reference and predicted text before and after punctuation and case-sensitivity restoration}
  \label{tab:longform-examples}
  \centering
  \vspace{-8pt}
  \scalebox{0.75}{
  \begin{tabular}{p{0.04\linewidth}p{0.6\linewidth}|p{0.6\linewidth}}
  \toprule
  Ref. & Original & LLM Punctuation and Case-Sensitivity Restoration \\
  \hline
  & he went toward the god and he made reverence and began to speak to him but apollo turned to admetus a face that was without joy what years of happiness have been mine o apollo through your friendship for me said admetus & 
  \textcolor{red}{H}e went toward the god and made reverence\textcolor{red}{,} and began to speak to him\textcolor{red}{. B}ut \textcolor{red}{A}pollo turned to \textcolor{red}{A}dmetus a face that was without joy\textcolor{red}{. 'W}hat years of happiness have been mine\textcolor{red}{, O A}pollo, through your friendship for me\textcolor{red}{?' } said Admetus\textcolor{red}{.}\\ 
  \toprule
  Pred. & OWSM v3.1 & OSWM v3.2 (This work)\\ 
  \hline
  & he went toward the god and he made reverence and began to speak to him but apollo turned to edmetus a face that was without joy what years of happiness have been mine o apollo through your friendship for me ned mettus & 
  \textcolor{red}{H}e went toward the \textcolor{red}{G}od\textcolor{red}{,} and he made reverence and began to speak to him\textcolor{red}{. B}ut \textcolor{red}{A}pollo turned to \textcolor{red}{A}dmettus a face that was without joy\textcolor{red}{. W}hat years of happiness have been mine\textcolor{red}{, O} \textcolor{red}{A}pollo, through your friendship for me\textcolor{red}{,} said \textcolor{red}{E}dmettus\textcolor{red}{.}\\ 
  \bottomrule
  \end{tabular}
  }
  \vspace{-10pt}
\end{table*}

\vspace{-3pt}
\subsection{Data Filtering}
\S2.1.1 indicates that the dataset can contain ill-aligned examples that can degrade model performance. We conjecture whether a model can achieve improved performance with reduced data volume but with higher data quality, i.e., by filtering out the low-quality data. 

To investigate the feasibility, our method starts with a proxy task.
We first conduct CTC greedy decoding using the existing OWSM v3.1 1B model \cite{owsm3.1} and compute the example-level character error rate (CER) using its label as the reference. The examples are then sorted by CER. The top-$k\%$ examples with the highest CER are considered of low quality and are discarded, where $k\%$ is a hyper-parameter\footnote{
Note OWSM v3.1 1B model is imperfect and this practice can wrongly discard some positive examples. However, we note this method is empirically effective and widely used in both dataset manufacturing \cite{gigaspeech} and active learning \cite{aishell_datadrop}.
}. 
The influence of $k\%$ is examined by training proxy models:
with each $k\%=$\{0\%, 5\%, 15\%, 25\%\}, a small proxy ASR model is trained based on the remaining 1-$k\%$ portion of data separately; the performance of the proxy model on a small validation set shows if discarding $k\%$ examples can provide improvement.

Considering the heterogeneity across the datasets, the proxy task is implemented separately on nearly every dataset. 
For multilingual ones, we combine every 5 languages with similar mean CER for one experiment group. 
The results of proxy tasks are in Table \ref{tab:proxy}.
As suggested in the table, in single-dataset scenarios, it is feasible to achieve performance improvement with reduced data volume but better data quality.
However, this observation is not consistent across datasets due to the heterogeneity described in \S2.1.1.
We observe that more than half (20 of 33) of these experiments achieve improvements with the data filtering method. This tendency encourages us to further investigate data filtering on the full-size experiment (see \S3.2).
We choose a unified $k\%=5\%$ based on Table \ref{tab:proxy}, assuming that a homogeneous protocol can reduce the heterogeneity across different datasets.

Additionally, we focus on LibriSpeech, GigaSpeech, and WenetSpeech due to the untranscribed clip issue in \S2.1.1
We additionally test $k\%=$\{35\%, 45\%\}, and found that larger $k\%$ can alleviate the deletion errors caused by the untranscribed clips.
Our proxy tasks suggest $k\%=$\{15\%, 35\%, 45\%\} provide the best performance on these datasets, respectively, so we discard the examples in these datasets accordingly. In total, we discard 27k-hour data, 15\% of the OWSM v3.1 training data\footnote{
For both ASR and ST data, we take ASR as the proxy task for uniformity and simplicity. For now, we do not consider the quality of $\textbf{y}^{\text{tgt}}$ and $\textbf{y}^{\text{prev}}$. For efficiency, each task only takes $N \times (1-k\%)$ randomly sampled utterances, where $N=50,000$. The proxy task is not conducted to several datasets due to their small volumes in each language, like FLEURS. 35\% is also applied to GigaST as it is derived from GigaSpeech.
}.

\begin{table}[h]
    \centering
    \caption{Proxy models' performance w.r.t. discarding ratio $k\%$}
    \vspace{-8pt}
    \scalebox{0.78}{
    \begin{tabular}{|l|c|c|c|c|c|c|c|c|}
        \hline
        \multirow{2}{*}{Dataset} & \multirow{2}{*}{Languages} & CTC &\multicolumn{4}{c|}{Proxy Task CER/WER\%}  \\ 
        \cline{4-7}
        && CER\% & k=25\% & k=15\% & k=5\% & k=0\% \\
        \hline
        aida. \cite{aidatatang} & zho & \phantom{0}8.9  &11.9 & 11.0 & \cellcolor{blue!25}10.0 & 11.2 \\
        \hline
        AISH. \cite{aishell-corpus} & zho & \phantom{0}8.1  & \phantom{0}8.5 & \phantom{0}7.8 & \phantom{0}7.0 & \cellcolor{blue!25}\phantom{0}6.6 \\
        \hline
        ami \cite{ami-corpus}& eng & 46.2  &22.5 & 22.1 & 21.2 & \cellcolor{blue!25}21.1 \\
        \hline
        \multirow{2}{*}{babel \cite{babel}} & tgl etc. & 54.7  & \cellcolor{blue!25}68.5 & 69.2 & 71.7 & 75.9 \\
          & pus etc. & 65.6 & 53.4 & 48.5 & \cellcolor{blue!25}47.3 & 48.1 \\
        \hline
        \multirow{11}{*}{CV\cite{commonvoice}} & bel etc. & \phantom{0}5.1  & 41.7 & 39.3 & \cellcolor{blue!25}35.6 & 36.5 \\ 
        & rus etc. & \phantom{0}9.9  & 37.9 & 38.0 & \cellcolor{blue!25}33.3 & 35.5 \\ 
        & nld etc. & 12.8 & 54.3 & 52.7 & \cellcolor{blue!25}48.9 & 50.2 \\
        & fas etc. & 16.2 & 56.9 & 54.8 & 52.3 & \cellcolor{blue!25}49.3 \\
        & pol etc. & 18.8 & 43.3 & 38.3 & 36.2 & \cellcolor{blue!25}34.6 \\
        & ind etc. & 22.6 & 53.5 & 53.2 & 49.9 & \cellcolor{blue!25}49.0 \\
        & uig etc. & 24.9 & 43.4 & 38.3 & 36.2 & \cellcolor{blue!25}34.6 \\
        & ukr etc. & 27.1 & 37.6 & 37.0 & \cellcolor{blue!25}33.2 & 34.6 \\
        & kmr etc. & 32.0 & 56.9 & 57.6 & 51.9 & \cellcolor{blue!25}50.2 \\
        & urd etc. & 38.7 & 53.0 & 49.1 & \cellcolor{blue!25}46.9 & 50.4 \\
        & lav etc. & 58.7 & 42.3 & 40.0 & \cellcolor{blue!25}39.9 & 41.1 \\
        \hline
        \multirow{2}{*}{CoVo. \cite{covost2}} & rus etc. & \phantom{0}6.9  & 45.8 & 43.9 & 42.7 & \cellcolor{blue!25}41.0 \\
        & fas etc. & 12.7  & 55.4 & 55.8 & \cellcolor{blue!25}47.9 & 50.9 \\
        \hline
        Fisher. \cite{fisher-callhome}& spa & 22.3  & \cellcolor{blue!25}22.4 & 22.6 & 23.0 & 22.6 \\
        \hline 
        kspo. \cite{ksponspeech} & kor & 16.5 & 14.5 & 13.2 & 12.5\cellcolor{blue!25} & 13.1 \\
        \hline
        magic. \cite{magicdata} & zho & \phantom{0}8.0 & 12.4 & 11.2 & 10.2\cellcolor{blue!25} & 11.1 \\
        \hline
        \multirow{2}{*}{MLS \cite{pratap2020mls}} & eng etc. & \phantom{0}5.7  & 33.2 & 32.0 & 31.4\cellcolor{blue!25} & 31.5 \\
        & fra etc.  & 11.6  & 21.6 & 21.7 & 20.7 & 20.5\cellcolor{blue!25} \\
        \hline
        must. \cite{must-c}& eng & \phantom{0}7.6  & 19.4 & 19.0 & \cellcolor{blue!25}18.4 & 18.4 \\
        \hline
        Googlei18n & sun etc. & \phantom{0}9.5  & 56.8 & 47.0 & 46.0 & \cellcolor{blue!25}30.7 \\
        \hline
        Reazo. \cite{reazonspeech}& jpn & 17.1  & 32.9 & 21.2 & \cellcolor{blue!25}19.4 & 21.0 \\
        \hline
        Russi. \cite{ru-open-stt}& rus & \phantom{0}8.0  & 35.2 & 34.5 & \cellcolor{blue!25}31.2 & 33.3 \\
        \hline
        SPGI. \cite{spgispeech} & eng & \phantom{0}2.7  & 10.9 & 10.1 & 10.5 & \cellcolor{blue!25}\phantom{0}9.2 \\
        \hline
        swbd \cite{swbd-corpus}& eng & 11.4  & 17.6 & 16.2 & 16.3 & \cellcolor{blue!25}15.0 \\
        \hline
        TEDL. \cite{tedlium3}& eng & \phantom{0}9.6  & \phantom{0}9.9 &\cellcolor{blue!25}\phantom{0}9.8 & 10.0 & \phantom{0}9.9 \\
        \hline
        VCTK \cite{vctk} & eng  & \phantom{0}3.0  & \phantom{0}9.0 & \phantom{0}8.1 & \phantom{0}4.1 & \cellcolor{blue!25}\phantom{0}3.9 \\
        \hline
        \multirow{2}{*}{VoxP. \cite{voxpopuli}} & eng etc. & 20.3 & 29.2 & \cellcolor{blue!25}28.2 & 28.4 & 28.6\\
                            & fra etc. & 25.4 & 31.9 & 31.8 & \cellcolor{blue!25}31.5 & 32.3 \\
        \hline\hline
        Count & & &  2 & 2 & \cellcolor{blue!25}16 & 13 \\
        \hline
    \end{tabular}}
    \label{tab:proxy}
\end{table}

\subsection{Punctuation and Case-Sensitivity Restoration}

Given the issue in \S2.1.2, this work restores the punctuation and case sensitivity in the training data using LLMs, specifically with the zero-shot prompt approach. An example is in Table \ref{tab:longform-examples}.
We find the English prompt below works well. The prompt is translated into other 8 languages for corresponding use cases\footnote{
8 languages: zho, deu, fra, spa, ita, nld, por, pol.}. 

\begin{tcolorbox}[colback=white,colframe=black, boxsep=0pt, left=2pt, right=2pt, top=2pt, bottom=2pt]
\footnotesize
For the given $<$language$>$ sentence, restore the upper-case characters (if applicable) and add punctuation WITHOUT CHANGING ANY WORDS. Answer in $<$language$>$ without any explanation. Here is the sentence: $<$input$>$. Here is the output:
\end{tcolorbox}


The stability of LLMs' outputs varies, occasionally altering the original text. Surprisingly, we find the capitalized phrase \textit{WITHOUT CHANGING ANY WORDS} in the prompt can effectively reduce this behavior. 
Next, to avoid unnecessary alterations, we compare the LLM output to the original text, only accepting (1) substitutions in casing and punctuation, and (2) punctuation insertions. 
We will not modify the text if the LLM output greatly differs from the original text (WER$>$30\% after the above changes are applied).
We only use LLMs to process $\mathbf{y}^{\text{tgt}}$ and then revise $\textbf{y}^{\text{src}}$ and $\textbf{y}^{\text{prev}}$ accordingly to reserve consistency of each example. 
To ensure reproducibility, the open-sourced LLM Mistral-7B-Instruct-v0.1 \cite{jiang2023mistral} is adopted. As diversity is not needed in this process, we use greedy search for LLM inference to exclude randomness.

\vspace{-5pt}
\section{Experiments}

\subsection{Experimental Setup}
\textbf{Proxy Models: }Same as \cite{owsm3.1}, the proxy tasks are implemented with the hybrid CTC/Attention framework \cite{hca}. 
We restrict the model parameters to around 20M for efficiency. All proxy models are updated for 100k steps. 
For each dataset, all setups are kept the same, except the training data. \\
\textbf{OWSM v3.2 models: } The OWSM v3.2 intentionally inherits the configurations in OWSM v3.1, except that the training data of OWSM v3.2 has experienced data filtering and punctuation and case-sensitivity restoration (\S2.2 \& \S2.3).
Specifically, our model adopts the same architecture and optimization strategy as OWSM v3.1-small, which contains 367M trainable parameters featured by E-Branchformer \cite{ebf}. 
The model is trained with the ESPnet \cite{espnet}, using 16 A100 40G GPUs for 9 days.
\\
\textbf{Evaluation: }All benchmarks included in \cite{owsm3.1} are also reported in this study. 
We additionally splice the short clips in Librispeech Test-\{Clean, Other\}, GigaSpeech Test, and WenetSpeech Test-Net, and then build a long-form subset without untranscribed clips for each test set. These subsets will be used to evaluate the model's long-form performance (\S3.3).
To verify the effectiveness of punctuation and case-sensitivity restoration, we additionally report the CER/WER\% that take punctuation and case-sensitivity into consideration (pc-CER/WER\%). The perplexity reported by another LLM without instruct fine-tuning, MPT-7B\footnote{https://huggingface.co/mosaicml/mpt-7b}, is used as an indicator of alignment with written text (\S3.4). All other evaluation setups follow \cite{owsm3.1}. In all our tables, all language-IDs follow ISO-639-3 standards.

\subsection{Main Results}
\begin{table}[t]
    \centering
    \caption{Main results of OWSM v3.2 compared with OWSM v3.1. 
    All setups are the same except the training data. 
    Performance measured by CER/WER\% for ASR and case-insensitive BLEU for ST.
    }
    \vspace{-5pt}
    \scalebox{0.72}{
    \begin{tabular}{|ccc|ccc|ccc|}
    \hline
    \multicolumn{3}{|c}{English ASR} & \multicolumn{3}{|c}{Multilingual ASR} & \multicolumn{3}{|c|}{X-eng ST \cite{covost2}}\\
    \hline
    Dataset & v3.1 & v3.2 &  Dataset & v3.1 & v3.2 & Lang. & v3.1 & v3.2 \\
    \hline
    CV     & 14.3 & \bf{13.9} & MLS-spa & 10.8 & \bf{10.5}& deu & 15.1 & \bf{15.5} \\
    FLEURS & 10.3 & \bf{10.1} & MLS-fra & 14.1 & 14.1& spa & 19.3 & \bf{20.3} \\
    LS Clean & \phantom{0}2.5 & \phantom{0}2.5 &  MLS-deu & 12.4 & \bf{12.1} & fra & 20.3 & \bf{21.6}\\
    LS Other & \phantom{0}\bf{5.8} & \phantom{0}6.2 & MLS-nld & 19.7 & \bf{19.5} & cat & 16.2 & \bf{17.1} \\
    MLS & \phantom{0}8.1 & \phantom{0}\bf{7.9} & MLS-ita & 21.8 & \bf{21.4} & & &\\
    SWBD & 17.4 & 17.4 &    MLS-por & 26.7 & \bf{25.9} & & & \\
    TEDLIUM & \phantom{0}\bf{5.0} & \phantom{0}5.4 & MLS-pol & \bf{28.5} & 31.7 & & & \\
    VoxPopuli & \phantom{0}9.1 & \phantom{0}\bf{9.0} & AISHELL & \phantom{0}\bf{7.5} & \phantom{0}8.2 & & & \\
    WSJ \cite{wsj} &  \phantom{0}\bf{3.8}  & \phantom{0}4.0  &   Kspon. clean & 17.2 & \bf{16.7} & & & \\ 
    & & & Kspon. other & 18.9 & \bf{18.5}  &&&\\ 
    & & & Reazon.   & \phantom{0}8.5 & \phantom{0}8.5 & & &\\
    \hline
    Average & \phantom{0}8.5 & \phantom{0}8.5 & & \bf{16.9} & 17.0 & & 17.7 & \bf{18.6} \\ 
    \hline\hline
    \multicolumn{9}{|c|}{eng-X ST \cite{covost2}} \\ 
    \hline
    Lang. & v3.1 & v3.2 &  Lang. & v3.1 & v3.2 & Lang. & v3.1 & v3.2 \\
    \hline
    deu &\bf{22.8} & 22.1 & cat &15.9& \bf{17.5} & zho &26.7& \bf{30.4} \\
    fas & \phantom{0}7.7 & \phantom{0}\bf{8.7}   & eos & \phantom{0}5.8 & \phantom{0}\bf{6.8} & mon &\phantom{0}3.3& \phantom{0}\bf{4.1} \\
    tur &\phantom{0}4.8 & \phantom{0}\bf{5.6}   & ara &\phantom{0}5.1& \phantom{0}\bf{5.8} & see   &16.6 & \bf{18.2} \\
    lav &\phantom{0}4.4 & \phantom{0}\bf{5.5}   & slv &\phantom{0}5.7 & \phantom{0}\bf{7.1} & tam  &\phantom{0}0.0 & \phantom{0}0.0  \\
    jpn &16.4 & \bf{16.8} & bdl &12.4 & \bf{15.2} & rmw &11.6& \bf{13.8} \\
    \hline 
    Average &10.6& \bf{11.8}&&&& &&\\
    \hline
    \end{tabular}}
    \vspace{-12pt}
    \label{tab:main_results}
\end{table}
Table \ref{tab:main_results} presents the main results of OWSM v3.2 compared with the OWSM v3.1 baseline to show the impact of data filtering.
Even with 15\% less data, OWSM v3.2 outperforms OWSM v3.1 consistently on ST tasks and achieves comparable performance on ASR benchmarks.
This observation partially supports our motivation to improve model performance with less data volume but better data quality. 
Specifically, the improvement in ST benchmarks implies that the ST tasks benefit more from the data quality improvement than ASR in this ASR-ST joint training scheme.
Aligned with Table \ref{tab:proxy}, the mixed ASR results in Table \ref{tab:main_results} also suggest the data heterogeneity persists in the full-size training. Though performance improvement is not achieved on ASR, our investigation implies that there can be considerable redundancy in the original data of OWSM v3.1\footnote{
The performance change shown in Table \ref{tab:main_results} should be more attributed to data filtering: (1) all metrics are calculated without punctuation and case-sensitivity; (2) our further small-scale experiments with data filtering only also show similar performance improvement.
}.

\subsection{Long-Form Results}
\begin{table}[t]
    \centering
    \caption{Comparison on long-form test sets. k\% means k\% training data is discarded in OWSM v3.2 but not in OWSM v3.1. Numbers in parentheses specify the deletion error rate. }
    \scalebox{0.85}{
    \begin{tabular}{|c|c|ccc|}
         \hline
         \multirow{2}{*}{Dataset} & \multirow{2}{*}{k\%} & \multicolumn{3}{c|}{CER/WER\%} \\
         \cline{3-5} 
         
         \cline{3-5}
         & & v3.1 & v3.2 & Relative Reduction \\ 
         \hline
         \multicolumn{5}{|c|}{Short-Clip}  \\ 
         \hline
         LS Clean & \multirow{2}{*}{15\%} & \phantom{0}2.5 (\textbf{0.3}) & \phantom{0}2.5 (0.4) & \phantom{0}0.0\% (+33.3\%) \\
         LS Other & & \phantom{0}\textbf{5.8} (0.6) & \phantom{0}6.2 (0.6)  &+6.8\% \phantom{00}(0.0\%)\\
         Giga. Test & 35\% & 12.7 (3.8) & \bf{12.0} (2.9) & -5.5\% (-23.6\%)\\
         Wenet. Net & 45\% & \bf{18.9} (4.4) &  19.8 (4.0)  & +4.7\% \phantom{0}(-9.1\%)\\
         \hline
         \multicolumn{5}{|c|}{Long-Form}  \\ 
         \hline
         LS Clean & \multirow{2}{*}{15\%} & \phantom{0}2.3 (0.5) & \phantom{0}\bf{2.1} (0.2) &  \phantom{0}-8.6\% (-60.0\%) \\
         LS Other & & \phantom{0}5.4 (1.2) & \phantom{0}\bf{4.8} (0.4) &  -11.1\% (-66.6\%)\\
         Giga. Test & 35\% & 17.8 (9.3)  & \bf{13.9} (3.7) & -21.9\% (-61.2\%)\\
         Wenet. Net & 45\% & 16.4 (5.1) &  \bf{15.7} (2.4)  & \phantom{0}-4.2\% (-52.9\%)\\
         \hline
    \end{tabular}}
    \vspace{-10pt}
    \label{tab:longform_results}
\end{table}

As in \S2.1.1, the untranscribed clip issue leads to increased deletion errors and then worsens the total performance.
Table \ref{tab:longform_results} shows how our data filtering method alleviates this issue. 
As suggested in the table, OWSM v3.2 consistently outperforms OWSM v3.1 in terms of long-form performance, even though a considerable portion of its training data has been filtered out.
Additionally, the reduction in CER/WER\% is specifically proportional to the deletion error reduction, which implies the improvement is mainly attributed to the alleviation of the untranscribed clip issue.
In terms of the short clip scenario, the impact of data filtering is neutral (LibriSpeech) or positive (GigaSpeech) but is negative on WenetSpeech. On WenetSpeech, although the deletion errors are still reduced in OWSM v3.2, it makes more substitution and insertion errors due to the greatly reduced training data volume (k\%=45\%).

\subsection{Punctuation and Case-Sensitivity Restoration Results}
OWSM v3.2 achieves tied ASR results with OWSM v3.1 in Table \ref{tab:main_results}. As CER/WER\% in Table \ref{tab:main_results} does not consider punctuation and case-sensitivity, it shows that restoring punctuation and case-sensitivity with LLMs will not degrade model performance with conventional evaluation metrics.

Table \ref{tab:punc} shows that the output of OWSM v3.2 is more aligned with written language: in all comparisons, OWSM v3.2  outperforms OWSM v3.1 in perplexity\footnote{
    On LS Clean, the perplexity from v3.2 is even better than the oracle.
}.
Additionally, for all English test sets, OWSM v3.2 outperforms OWSM v3.1 in terms of pc-WER. 
Our method improves pc-WER on FLEURS, which originally contains punctuation and case-sensitivity. The improvement achieved on FLEURS suggests the punctuated and case-sensitive text output by OWSM v3.2 is better aligned with real scenarios.
The pc-WER of OWSM v3.2 is worse than OWSM v3.1 on MLS Spanish and French, but it is mainly attributed to the poor reference generated by LLM\footnote{
The LLM \cite{jiang2023mistral} is not designed for multilingual usage. We find the non-English reference text generated by the LLM is poor; e.g., for Spanish and French, the first character is often not capitalized.
}. 

\begin{table}[h]
    \centering
    \vspace{-5pt}
    \caption{Effectiveness of punctuation and case-sensitivity restoration. * suggests the reference is generated with LLM. Oracle is obtained by reference text.}
    \scalebox{0.95}{
    \begin{tabular}{|c|cc|cc||c|}
         \hline
         \multirow{2}{*}{Dataset}  & \multicolumn{2}{c|}{pc-WER\%} & \multicolumn{3}{c|}{LLM Perplexity} \\
         \cline{2-6}
         & v3.1 & v3.2 & v3.1 & v3.2 & Oracle \\ 
         \hline
         LS Clean* & 23.2 & \phantom{0}\bf{7.8} & 592.0 & \bf{100.7} & 127.7 \\
         MLS-eng* & 26.4 & \bf{15.1} &  157.5 & \phantom{0}\bf{85.4} & \phantom{0}64.9\\
         FLEURS-eng & 17.6 & \bf{15.9} & \phantom{0}85.6 & \phantom{0}\bf{69.6} & \phantom{0}28.6\\
         \hline
         MLS-spa* & \bf{18.5} & 19.1 & \phantom{0}63.2& \phantom{0}\bf{46.6} & \phantom{0}39.3\\
         MLS-fra* & \bf{21.0} & 23.3 & \phantom{0}50.3& \phantom{0}\bf{33.1} & \phantom{0}27.8\\
         FLEURS-spa & 18.6 & \bf{17.7} & \phantom{0}42.3 & \phantom{0}\bf{32.3} & \phantom{0}16.3 \\
         FLEURS-fra & 24.7& \bf{23.3} & \phantom{0}45.1 & \phantom{0}\bf{36.2} & \phantom{0}13.4 \\
         \hline
    \end{tabular}}
    \vspace{-5pt}
    \label{tab:punc}
\end{table}

\section{Conclusion}
This work presents Whisper-style Speech Model (OWSM) v3.2, which is distinctively designed to address the heterogeneity introduced by the diverse data compositions. 
By utilizing proxy tasks for data filtering and leveraging Large Language Models (LLMs) for punctuation and case-sensitivity restoration, the model is optimized for ST performance and output readability.

\clearpage
\section{Acknowledgements}
Some experiments of this work used the Bridges2 system at PSC and Delta system at NCSA through allocation CIS210014 from the Advanced Cyberinfrastructure Coordination Ecosystem: Services \& Support (ACCESS) program, which is supported by National Science Foundation grants \#2138259, \#2138286, \#2138307, \#2137603, and \#2138296. We also gratefully acknowledge the support of NVIDIA Corporation with the donation of the Quadro RTX 8000 GPUs used for this research.

\bibliographystyle{IEEEtran}
\bibliography{mybib}

\begin{thebibliography}{10}
\providecommand{\url}[1]{#1}
\csname url@samestyle\endcsname
\providecommand{\newblock}{\relax}
\providecommand{\bibinfo}[2]{#2}
\providecommand{\BIBentrySTDinterwordspacing}{\spaceskip=0pt\relax}
\providecommand{\BIBentryALTinterwordstretchfactor}{4}
\providecommand{\BIBentryALTinterwordspacing}{\spaceskip=\fontdimen2\font plus
\BIBentryALTinterwordstretchfactor\fontdimen3\font minus \fontdimen4\font\relax}
\providecommand{\BIBforeignlanguage}[2]{{%
\expandafter\ifx\csname l@#1\endcsname\relax
\typeout{** WARNING: IEEEtran.bst: No hyphenation pattern has been}%
\typeout{** loaded for the language `#1'. Using the pattern for}%
\typeout{** the default language instead.}%
\else
\language=\csname l@#1\endcsname
\fi
#2}}
\providecommand{\BIBdecl}{\relax}
\BIBdecl

\bibitem{asr_survey2}
R.~Prabhavalkar \emph{et~al.}, ``End-to-end speech recognition: A survey,'' \emph{IEEE/ACM TASLP}, vol.~32, pp. 325--351, 2024.

\bibitem{st_survey2}
M.~Agarwal \emph{et~al.}, ``Findings of the {IWSLT} 2023 evaluation campaign,'' in \emph{IWSLT}, 2023.

\bibitem{google-usm}
Y.~Zhang \emph{et~al.}, ``Google {USM}: Scaling automatic speech recognition beyond 100 languages,'' \emph{arXiv preprint arXiv:2303.01037}, 2023.

\bibitem{seamless}
L.~Barrault \emph{et~al.}, ``Seamless: Multilingual expressive and streaming speech translation,'' \emph{arXiv preprint arXiv:2312.05187}, 2023.

\bibitem{seamlessm4t}
L.~Barrault,  \emph{et~al.}, ``Seamlessm4t-massively multilingual \& multimodal machine translation,'' \emph{arXiv preprint arXiv:2308.11596}, 2023.

\bibitem{pratap2023scaling}
V.~Pratap \emph{et~al.}, ``Scaling speech technology to 1,000+ languages,'' \emph{JMLR}, 2024.

\bibitem{whisper}
A.~Radford \emph{et~al.}, ``Robust speech recognition via large-scale weak supervision,'' in \emph{ICML}, 2023.

\bibitem{owsm}
Y.~Peng \emph{et~al.}, ``{Reproducing Whisper-Style Training Using an Open-Source Toolkit and Publicly Available Data},'' in \emph{Proc. ASRU}, 2023.

\bibitem{owsm3.1}
P.~Yifan \emph{et~al.}, ``Owsm v3.1: Better and faster open whisper-style speech models based on e-branchformer,'' \emph{Interspeech}, 2024.

\bibitem{owsm_ctc}
Y.~Peng \emph{et~al.}, ``{OWSM-CTC}: An open encoder-only speech foundation model for speech recognition, translation, and language identification,'' \emph{ACL}, 2024.

\bibitem{gigaspeech}
G.~Chen \emph{et~al.}, ``{GigaSpeech: An Evolving, Multi-Domain ASR Corpus with 10,000 Hours of Transcribed Audio},'' in \emph{Interspeech}, 2021.

\bibitem{wenetspeech}
B.~Zhang \emph{et~al.}, ``Wenetspeech: A 10000+ hours multi-domain mandarin corpus for speech recognition,'' in \emph{ICASSP}, 2022.

\bibitem{takamichi2021jtubespeech}
S.~Takamichi \emph{et~al.}, ``Jtubespeech: corpus of japanese speech collected from youtube for speech recognition and speaker verification,'' \emph{arXiv preprint arXiv:2112.09323}, 2021.

\bibitem{librispeech-corpus}
V.~Panayotov \emph{et~al.}, ``{Librispeech: An ASR corpus based on public domain audio books},'' in \emph{ICASSP}, 2015.

\bibitem{lu22_interspeech}
Z.~Lu \emph{et~al.}, ``{Unsupervised Data Selection via Discrete Speech Representation for ASR},'' in \emph{Interspeech}, 2022.

\bibitem{shef_contrastive}
C.~Park \emph{et~al.}, ``Unsupervised data selection for speech recognition with contrastive loss ratios,'' in \emph{ICASSP}, 2022.

\bibitem{aishell_datadrop}
Y.~Chen \emph{et~al.}, ``Improving noisy student training on non-target domain data for automatic speech recognition,'' in \emph{ICASSP}, 2023.

\bibitem{9383577}
A.-L. Georgescu \emph{et~al.}, ``Data-filtering methods for self-training of automatic speech recognition systems,'' in \emph{SLT}, 2021.

\bibitem{zhang21ja_interspeech}
Y.~Zhang \emph{et~al.}, ``{NeMo (Inverse) Text Normalization: From Development to Production},'' in \emph{Interspeech}, 2021.

\bibitem{sproat2016rnn}
R.~Sproat and N.~Jaitly, ``An rnn model of text normalization,'' \emph{Interspeech}, 2017.

\bibitem{paul22_interspeech}
D.~Paul \emph{et~al.}, ``{Improving Data Driven Inverse Text Normalization using Data Augmentation and Machine Translation},'' in \emph{Interspeech}, 2022.

\bibitem{ling2023adapting}
S.~Ling \emph{et~al.}, ``Adapting large language model with speech for fully formatted end-to-end speech recognition,'' in \emph{ICASSP}, 2024.

\bibitem{10389653}
Y.~Huang \emph{et~al.}, ``Multi transcription-style speech transcription using attention-based encoder-decoder model,'' in \emph{ASRU}, 2023.

\bibitem{aidatatang}
``{aidatatang\_200zh, a free Chinese Mandarin speech corpus by Beijing DataTang Technology Co., Ltd}.''

\bibitem{aishell-corpus}
H.~Bu \emph{et~al.}, ``{AISHELL-1: An open-source Mandarin speech corpus and a speech recognition baseline},'' in \emph{O-COCOSDA}, 2017.

\bibitem{ami-corpus}
J.~Carletta, ``{Unleashing the killer corpus: experiences in creating the multi-everything AMI Meeting Corpus},'' \emph{Lang. Res. Eval.}, vol.~41, pp. 181--190, 2007.

\bibitem{babel}
``The babel program: https://www.iarpa.gov/index.php/research-programs/babel.''

\bibitem{commonvoice}
R.~Ardila \emph{et~al.}, ``Common voice: A massively-multilingual speech corpus,'' in \emph{Proceedings of the Twelfth Language Resources and Evaluation Conference}, 2020.

\bibitem{covost2}
C.~Wang \emph{et~al.}, ``{CoVoST 2 and Massively Multilingual Speech Translation},'' in \emph{Interspeech}, 2021.

\bibitem{fisher-callhome}
M.~Post \emph{et~al.}, ``Improved speech-to-text translation with the fisher and callhome {S}panish-{E}nglish speech translation corpus,'' in \emph{IWSLT}, 2013.

\bibitem{FLEURS}
A.~Conneau \emph{et~al.}, ``{FLEURS: Few-Shot Learning Evaluation of Universal Representations of Speech},'' in \emph{SLT}, 2022.

\bibitem{gigast}
R.~Ye \emph{et~al.}, ``{GigaST: A 10,000-hour Pseudo Speech Translation Corpus},'' in \emph{Interspeech}, 2023.

\bibitem{ksponspeech}
J.-U. Bang \emph{et~al.}, ``Ksponspeech: Korean spontaneous speech corpus for automatic speech recognition,'' \emph{Applied Sciences}, vol.~10, no.~19, p. 6936, 2020.

\bibitem{magicdata}
Z.~Yang \emph{et~al.}, ``Open source magicdata-ramc: A rich annotated mandarin conversational (ramc) speech dataset,'' \emph{arXiv:2203.16844}, 2022.

\bibitem{pratap2020mls}
V.~Pratap \emph{et~al.}, ``{MLS}: A large-scale multilingual dataset for speech research,'' in \emph{Interspeech}, 2020.

\bibitem{must-c}
R.~Cattoni \emph{et~al.}, ``Mu{ST-C}: A multilingual corpus for end-to-end speech translation,'' \emph{Computer speech \& language}, vol.~66, p. 101155, 2021.

\bibitem{reazonspeech}
Y.~Yin, D.~Mori \emph{et~al.}, ``{ReazonSpeech: A Free and Massive Corpus for Japanese ASR},'' 2023.

\bibitem{ru-open-stt}
A.~Slizhikova \emph{et~al.}, ``{Russian Open Speech To Text (STT/ASR) Dataset},'' 2020.

\bibitem{spgispeech}
P.~K. O’Neill \emph{et~al.}, ``{SPGISpeech: 5,000 Hours of Transcribed Financial Audio for Fully Formatted End-to-End Speech Recognition},'' in \emph{Interspeech}, 2021.

\bibitem{swbd-corpus}
J.~Godfrey \emph{et~al.}, ``{SWITCHBOARD: telephone speech corpus for research and development},'' in \emph{ICASSP}, 1992.

\bibitem{tedlium3}
F.~Hernandez \emph{et~al.}, ``{TED-LIUM} 3: Twice as much data and corpus repartition for experiments on speaker adaptation,'' in \emph{Speech \& Computer}, 2018, pp. 198--208.

\bibitem{vctk}
J.~Yamagishi \emph{et~al.}, ``{CSTR VCTK Corpus: English Multi-speaker Corpus for CSTR Voice Cloning Toolkit},'' 2019.

\bibitem{voxforge}
``{VoxForge: http://www.voxforge.org/}.''

\bibitem{voxpopuli}
C.~Wang \emph{et~al.}, ``{VoxPopuli: A Large-Scale Multilingual Speech Corpus for Representation Learning, Semi-Supervised Learning and Interpretation},'' in \emph{ACL}, 2021.

\bibitem{fox2023updated}
J.~D. Fox \emph{et~al.}, ``Updated corpora and benchmarks for long-form speech recognition,'' in \emph{ICASSP}, 2024.

\bibitem{jiang2023mistral}
A.~Q. Jiang \emph{et~al.}, ``Mistral 7b,'' \emph{arXiv preprint arXiv:2310.06825}, 2023.

\bibitem{hca}
S.~Watanabe \emph{et~al.}, ``Hybrid ctc/attention architecture for end-to-end speech recognition,'' \emph{IEEE JSTSP}, vol.~11, no.~8, pp. 1240--1253, 2017.

\bibitem{ebf}
K.~Kim \emph{et~al.}, ``E-branchformer: Branchformer with enhanced merging for speech recognition,'' in \emph{SLT}, 2023.

\bibitem{espnet}
S.~Watanabe \emph{et~al.}, ``{ESPnet: End-to-End Speech Processing Toolkit},'' in \emph{Interspeech}, 2018.

\bibitem{wsj}
D.~B. Paul and J.~Baker, ``{The design for the Wall Street Journal-based CSR corpus},'' in \emph{Proc. Workshop on Speech and Natural Language}, 1992.

\end{thebibliography}

\end{document}